\documentclass{article}
\usepackage{ijcai25}

\usepackage{times}
\usepackage{soul}
\usepackage{url}
\usepackage[hidelinks]{hyperref}
\usepackage[utf8]{inputenc}
\usepackage[small]{caption}
\usepackage{graphicx}
\usepackage{amsmath}
\usepackage{amsthm}
\usepackage{booktabs}
\usepackage{algorithm}
\usepackage{algorithmic}
\usepackage[switch]{lineno}
\usepackage{array}
\usepackage{amsfonts} 
\usepackage{amssymb}  
\usepackage[switch]{lineno}

\title{From Task-Specific Models to Unified Systems: A Review of Model Merging Approaches}

\author{
Wei Ruan$^1$
\and
Tianze Yang$^1$\and
Yifan Zhou$^1$\And
Tianming Liu$^1$\and
Jin Lu$^1$\\
\affiliations
$^1$School of Computing, University of Georgia, GA, USA\\
\emails
\{wei.ruan, tianze.yang, yifan.zhou1, tliu, jin.lu\}@uga.edu
}

\begin{document}

\maketitle

\begin{abstract}
Model merging has achieved significant success, with numerous innovative methods proposed to enhance capabilities by combining multiple models. However, challenges persist due to the lack of a unified framework for classification and systematic comparative analysis, leading to inconsistencies in terminologies and categorizations. Meanwhile, as an increasing number of fine-tuned models are publicly available, their original training data often remain inaccessible due to privacy concerns or intellectual property restrictions. This makes traditional multi-task learning based on shared training data impractical. In scenarios where direct access to training data is infeasible, merging model parameters to create a unified model with broad generalization across multiple domains becomes crucial, further underscoring the importance of model merging techniques. Despite the rapid progress in this field, a comprehensive taxonomy and survey summarizing recent advances and predicting future directions are still lacking. This paper addresses these gaps by establishing a new taxonomy of model merging methods, systematically comparing different approaches, and providing an overview of key developments. By offering a structured perspective on this evolving area, we aim to help newcomers quickly grasp the field's landscape and inspire further innovations.
\end{abstract}

\section{Introduction}
The increasing prevalence of open-source models trained on diverse tasks provides unprecedented opportunities to utilize pre-trained weights for various applications. However, access to the original training data is often restricted due to privacy concerns, proprietary limitations, or other constraints, posing significant challenges for tasks requiring cross-domain capabilities~\cite{jin2022dataless}. Model merging techniques address this issue by enabling the combination of model weights without relying on original data, thereby equipping models with the ability to handle multiple tasks effectively. Currently, model merging has emerged as a promising solution, enabling the combination of multiple models with similar architectures to harness complementary strengths. This approach not only enhances task-specific performance but also fosters greater adaptability across tasks~\cite{yang2024model,tam2024llm}.

Model merging provides several key advantages~\cite{yang2024model,yu2024language,zhao2024adamergex}. Firstly, it allows for the aggregation of knowledge across multiple models without requiring extensive retraining, thereby offering a more resource-efficient alternative to traditional fine-tuning and transfer learning. Additionally, model merging can mitigate issues like catastrophic forgetting and offers a pathway to create models that encapsulate the strengths of multiple training regimes. For instance, weight average~\cite{wortsman2022model,choshen2022fusing} and task arithmetic merging~\cite{ilharco2022editing} are widely adopted methods for retaining model capabilities while maintaining robustness across varied domains.

In recent years, model merging techniques have evolved from simple linear interpolation methods or weight averaging to more sophisticated approaches~\cite{yang2024model,sung2023empirical}. These include weight interference suppression, parameter freezing, and decoupling parameters for old and new tasks, allowing for the fine-tuning of specific model aspects while preserving core functionalities. There is also growing interest in integrating merging methods with Mixture of Experts (MoE) frameworks, where specialized "experts" are dynamically engaged based on the task requirements. These advancements underscore the potential of model merging as a versatile solution, capable of adapting to a range of tasks and minimizing resource demands.

\begin{figure}[htbp]
    \centering
    \includegraphics[width=\linewidth]{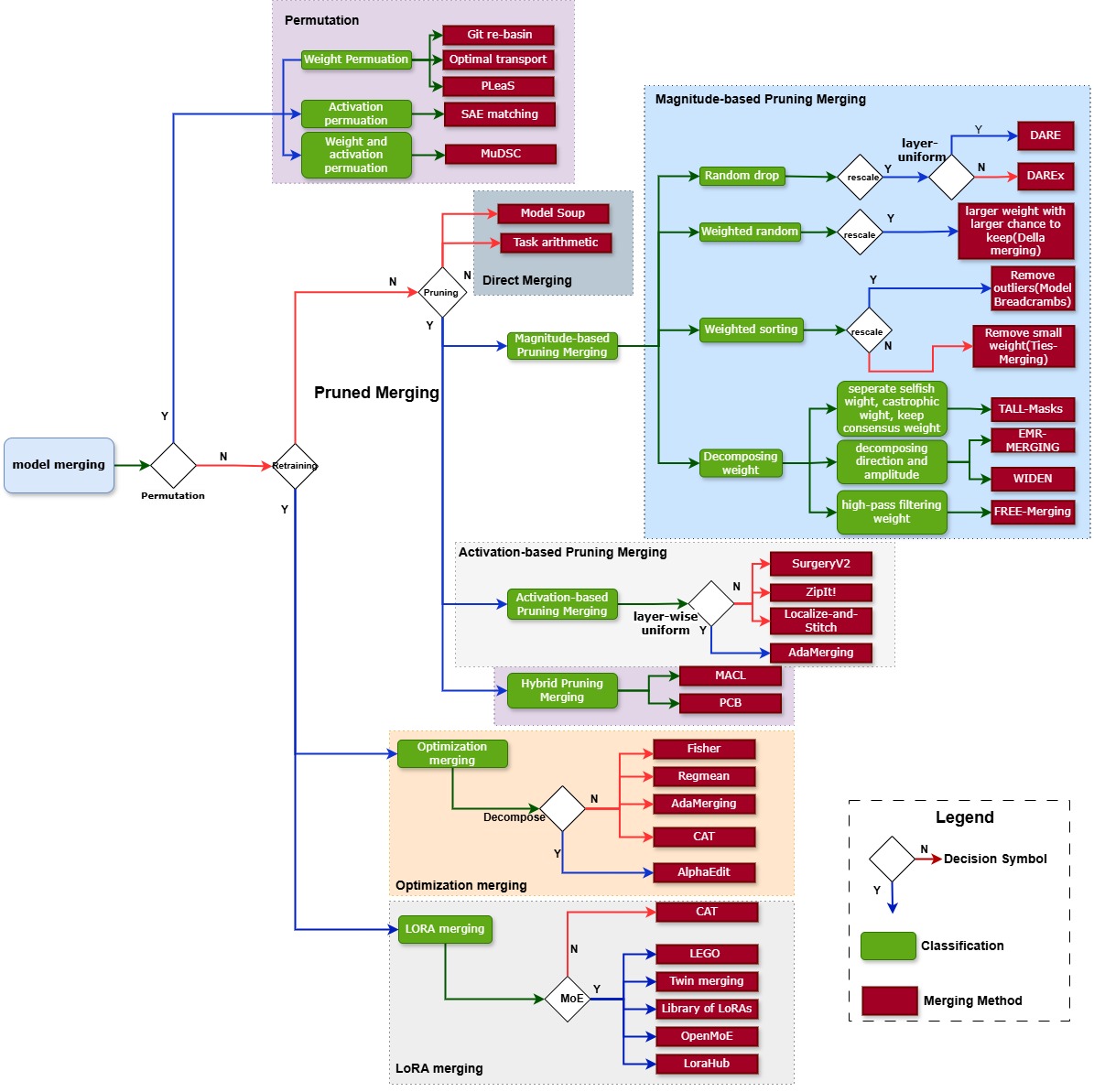}
    \caption{An overview of common model merging methods and their main distinctions.}
    \label{fig:merging-1}
\end{figure}

Although model merging has achieved remarkable results and numerous innovative methods have been developed, drawing increasing attention from researchers, this field is still in its early stages and faces several challenges. First, existing methods lack a unified framework for classification and comparative analysis. There is no comprehensive perspective to systematically understand and define model merging as a discipline, leaving inconsistencies in terminologies and categorizations. Second, the field lacks in-depth surveys that provide a thorough investigation of existing approaches. This absence makes it difficult for newcomers to quickly and comprehensively grasp the methodologies in this area. Moreover, there is a lack of systematic and forward-looking thinking about the future trends and potential development of this domain. This paper seeks to address these gaps by presenting a new taxonomy of model merging methods (Figure~\ref{fig:merging-1}). We categorize and analyze various approaches, examine the distinct challenges faced in model merging, and propose a comprehensive framework for advancing this research area . Our aim is to provide a foundational reference for researchers interested in model merging and to catalyze future innovations in this field, bridging the existing knowledge gaps and helping newcomers rapidly understand this domain.

\section{Classification of Model Merging}

\subsection{Permutation type}

Model merging was initially introduced to address challenges arising from Linear Mode Connectivity (LMC) barriers and Permutation Invariance~\cite{garipov2018loss}. LMC barriers refer to the obstacles encountered when attempting to linearly interpolate between two independently trained neural networks, which often results in suboptimal performance due to differences in their parameter spaces~\cite{zhou2023going,singh2024landscaping}. These barriers highlight the difficulty in aligning the solutions found by stochastic gradient descent (SGD), as they typically reside in separate loss basins~\cite{freeman2016topology}. To overcome LMC barriers, various approaches have been proposed, including post-hoc neuron alignment techniques, permutation symmetry reduction methods, and training-time alignment strategies, each aiming to improve the connectivity and compatibility of independently trained models~\cite{li2024training}.

In neural network training, independently trained models with the same architecture often exhibit unaligned weights, a phenomenon stemming from permutation invariance. This invariance implies that the order of neurons within a layer does not affect the overall functionality of the network. As a result, when two models trained separately are merged, their corresponding weights do not naturally align, making direct merging ineffective or suboptimal~\cite{ainsworth2022git,zhou2023going}.

Without alignment, merging such models can result in high loss barriers between their weights, which hinders Linear Mode Connectivity (LMC)~\cite{zhou2023going}. LMC refers to the existence of a smooth path in the loss landscape that connects two models. When weights are misaligned, naive merging can degrade model performance due to incompatibilities in the combined representations~\cite{liunveiling}.

Permutation-type model merging assumes that the models are trained on the same dataset but with different initializations and training parameters. Currently, several commonly used methods exist to align the weights before merging neural networks:

\textbf{Weight Matching:} Due to the permutation invariance property of neural networks, neurons within a layer can be arbitrarily reordered without altering the model's functionality. Weight Matching leverages this property by searching for an optimal permutation matrix that minimizes the L2 distance between the parameters of two models. This is achieved by aligning corresponding neurons in a way that ensures the functional similarity of the models is preserved~\cite{ito2024analysis}.This method solves a Linear Assignment Problem (LAP) to match the rows and columns of weight matrices between two models. It minimizes the difference between weights by finding the best permutation~\cite{ainsworth2022git,mena2018learning,cai2024neural,ito2024analysis}.Neural Functional Networks (NFNs)~\cite{zhou2024permutation}leverage the permutation invariance property of neural network weights to achieve significant advantages in weight-space processing tasks. By designing permutation-equivariant layers that respect the inherent symmetry of neural network weights, NFNs enable more accurate and efficient handling of tasks such as weight-based generalization prediction, classification of implicit neural representations (INRs), and weight-space style editing. This approach ensures that the arrangement of weights does not affect the network's output while capturing essential structural information, leading to improved performance across diverse applications.
PLeaS is a weight-based matching method for model merging that achieves excellent performance by aligning weights through permutation and optimization~\cite{nasery2024pleas}. However, its high computational cost limits its scalability to large models.

\textbf{Activation Matching:} Instead of focusing on weights, this method aligns neurons based on their activation patterns. It uses similarity in neuron activations to guide the alignment, ensuring functional equivalence~\cite{ainsworth2022git,entezari2021role,tatro2020optimizing,balagansky2024mechanistic}.

\textbf{Combination Matching:} This approach combines weight matching with activation matching. For example, Git Re-Basin~\cite{ainsworth2022git} leverages activation similarities and weight structure to align models effectively, addressing permutation symmetries during model merging. Other methods, such as those utilizing gradient-based alignment~\cite{li2024training} or optimal transport~\cite{mena2017sinkhorn}, further enhance matching by integrating activation dynamics and weight correlations. These approaches aim to achieve more robust feature alignment across layers, contributing to better model interpretability and performance in tasks requiring precise feature correspondence.

Among the many methods of combination matching, the MuDSC method is a relatively representative example.This approach focuses on model merging by first aligning the models based on a combination of weight and activation similarities, followed by the merging process.~\cite{xu2024training}. This dual-space alignment ensures that models are accurately matched in both representation spaces, addressing the challenge of merging models that were not pretrained with a common initialization. By overcoming this limitation, MuDSC enables the seamless integration of diverse pretrained models, eliminating the need for additional training or fine-tuning. This approach significantly enhances the flexibility and applicability of model merging across various architectures and tasks.

Permutation-based model merging is mainly used for the same task with different initializations and training parameters. After merging, it improves the model's generalization. However, since it is designed for the same dataset and task, its application is limited~\cite{zhouemergence,chen2023going}. That said, permutation-based methods have played a crucial role in the development of model merging techniques, especially in theoretical research.

\subsection{Direct Merging}
In the pretraining-finetuning paradigm, different model branches that share a common pretrained initialization exhibit specific linear properties. These properties, such as Cross-Task Linearity (CTL), allow for the linear interpolation of weights and corresponding feature spaces across tasks. This phenomenon enables effective model merging by ensuring consistency between parameter spaces and feature representations, even when the models are fine-tuned on different tasks. It provides a foundation for combining diverse task-specific models into a unified framework~\cite{zhouemergence}.

Direct Merging type is a relatively simple model merging method that avoids the complexities of resolving conflicts and interferences during the merging of model weights. This approach does not require retraining, significantly reducing computational costs. Moreover, many advanced model merging methods are built upon these straightforward foundational approaches, highlighting their importance in the field. Despite their simplicity, these foundational methods serve as the cornerstone for further innovations and advancements in model merging techniques. Several model merging methods fall into this category. The following sections introduce these approaches.

\textbf{Model Soup ( Weight Averaging ):} It is a method for merging fine-tuned models by averaging their weights, improving accuracy and robustness without additional training or inference costs. It works well when models share the same pretrained initialization but differ in hyperparameters~\cite{wortsman2022model}. Variants like uniform averaging and greedy soup selectively optimize performance. Model soup is efficient, flexible, and outperforms individual models, making it a robust alternative to ensembles for tasks like image and text classification. However, this method does not account for weight conflicts during model merging, which often results in performance that falls short of more sophisticated model merging approaches.

\textbf{Task arithmetic:} It is a method for editing pre-trained models by leveraging task vectors, which are computed as the difference between a model's fine-tuned weights and its pre-trained weights (\(\tau_t = \theta_{t} - \theta_0\)). These vectors encapsulate the changes needed for a model to perform a specific task and can be applied to modify model weights through simple operations like addition, subtraction, and analogy-based reasoning~\cite{ilharco2022editing}.

Key operations include task forgetting (negating task vectors to reduce performance on specific tasks, e.g., mitigating harmful behaviors like toxic text generation), task learning (adding task vectors to create multi-task models or enhance single-task performance), and task analogies (combining task vectors to improve performance on new tasks with limited or no labeled data). 
\begin{figure}[htbp]
    \centering
    \includegraphics[width=\linewidth]{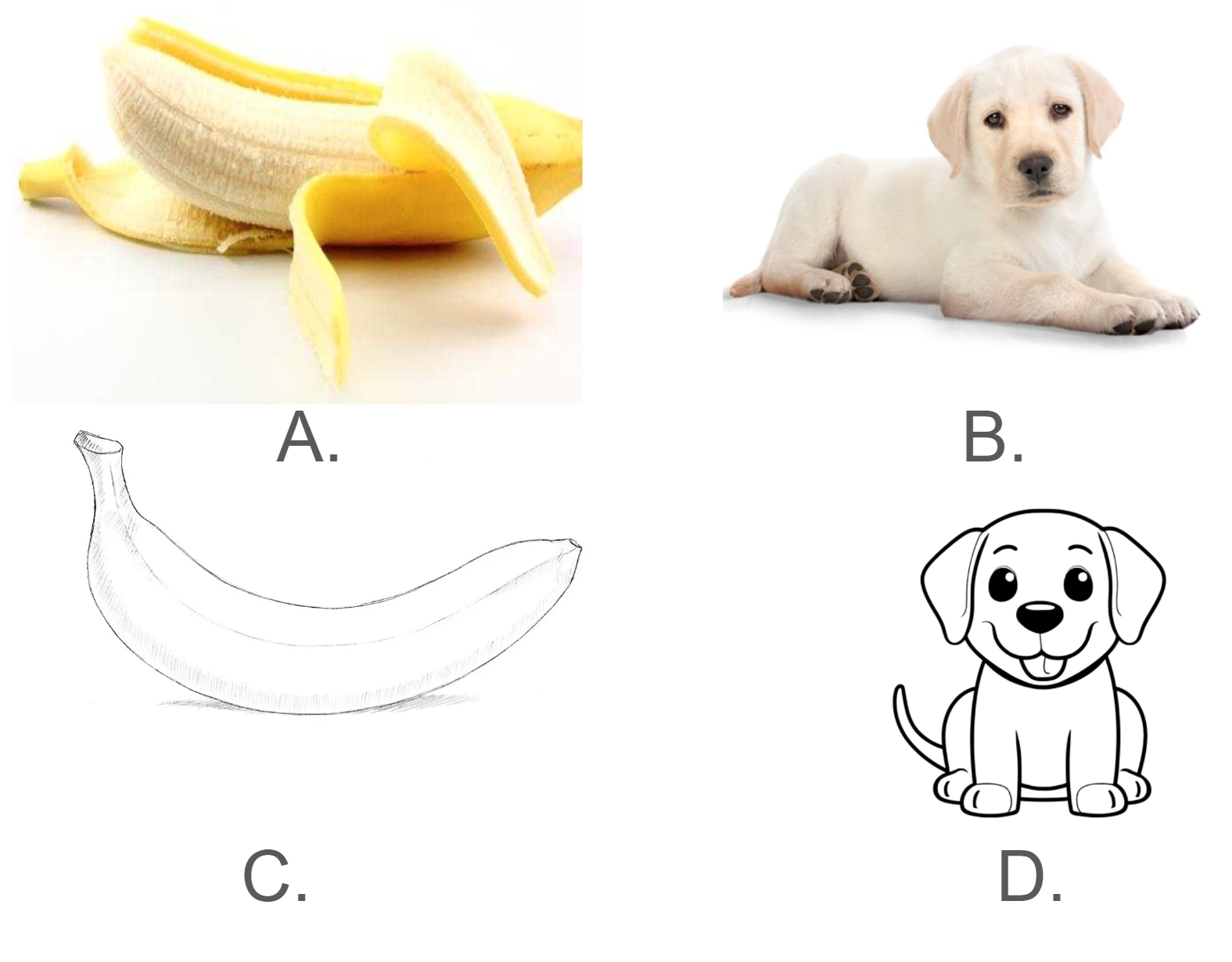} 
    \caption{The cartoon diagram illustrates the principle of task arithmetic.} 
    \label{fig:1} 
\end{figure}
\begin{equation}
\hat{\tau}_{\text{ dog cartoon}} = \tau_{\text{dog picture}} + \left( \tau_{\text{banana cartoon}} - \tau_{\text{banana picture}} \right)
\label{cartoon}
\end{equation}

To better illustrate the principle of task arithmetic, let us consider an example. Suppose we have original banana images and cartoon banana images, which have been used to train separate models. Similarly, we have original dog images that have been used to train another model. However, we lack cartoon dog images in the training dataset. To enhance the predictive accuracy for cartoon dog images, as shown in Equation~\ref{cartoon} and Figure~\ref{fig:1}, task arithmetic operations can be applied. This approach utilizes the relationships between tasks and models to improve performance on the target task, in this case, predicting cartoon dog images.

\subsection{Magnitude-based Pruning Merging}

In model merging, various types of interference can occur, leading to significant challenges~\cite{hoefler2021sparsity,jin2022dataless}. These interferences include conflicts between parameters, inconsistencies in feature representations, and task-related competition, which can result in degraded model performance or instability after merging. As a result, addressing interference before model merging has become a critical research focus~\cite{yadav2024ties}.

Based on the methods used to suppress interference, the approaches can be categorized into the following subtypes: random drop, weighted random drop, and decomposing weight. Below, we will discuss each of these types in detail. Specifically, this category can be further divided into the following subtypes.

\subsubsection{Random drop type}
The random drop method involves randomly replacing certain fine-tuned model weights with pre-trained model weights, effectively resetting those parameters to their original values. This approach increases the sparsity of the model, thereby reducing parameter conflicts during model merging. Furthermore, this method can be further subdivided into several specific variants. The following methods are the most well-known among models that use random drop.

\textbf{DARE(Drop And Rescale) :} This method focuses on reducing delta parameter redundancy before merging. DARE operates in two steps: it randomly drops a proportion of delta parameters (defined as the differences between fine-tuned and pre-trained model parameters) and rescales the remaining parameters to maintain their original magnitude (Table\ref{tab:methods}). By eliminating up to 90\% of redundant delta parameters without significantly impacting model performance, DARE mitigates interference during the merging process. Once sparsified by DARE, the models are merged using established methods like Task Arithmetic or TIES-Merging. This preprocessing step ensures a smoother merging process and often enhances the merged model's performance, particularly in large-scale language models~\cite{yu2024language,deep2024della}.

DAREx: The DARE (Drop And REscale) method, which randomly drops delta parameters and rescales the remaining ones, performs well at moderate drop rates but degrades significantly under extreme pruning, such as 99\%. To address this, DAREx introduces enhancements that adjust the rescaling factor for better verification and enforce sparsity constraints with 
$q > 1-p$. These improvements balance the mean and variance of delta parameters, mitigating performance loss and maintaining competitive results even under high pruning rates~\cite{deng2024dare}.

\subsubsection{Weighted random drop type}
In the weighted random drop approach to model merging, parameters are no longer dropped with equal probability. Instead, parameters with larger weights are less likely to be dropped, while those with smaller weights have a higher likelihood of being removed. This strategy aims to preserve important parameters, ensuring that the performance of the merged model is not significantly degraded, while simultaneously eliminating noise-induced large-weight parameters, thereby enhancing the generalization capability of the model. The following methods fall into this category.

\textbf{DELLA-Merging:} This method consists of three steps: Drop, Elect, and Fuse, designed to effectively merge multiple task-specific models~\cite{deep2024della}.

DELLA-Merging consists of three steps: Drop, Elect, and Fuse to merge task-specific models effectively.

Drop: MAGPRUNE prunes delta parameters based on their magnitude, assigns a dropout probability inversely proportional to their rank, and rescales the remaining parameters to compensate for the dropout effect.

Elect: Selects delta parameters that minimize interference by choosing those with consistent signs across models.

Fuse: Averages the selected parameters and scales them to obtain the final merged model.

By reducing interference while preserving task-specific information, DELLA outperforms methods like TIES and DARE.

\subsubsection{Weight sorting type}
As the name suggests, this method assumes that parameters with larger magnitudes have a greater impact on model performance and should be preserved. The approach involves sorting parameters by their absolute values, retaining those with larger magnitudes while dropping those with smaller absolute values. This helps reduce parameter interference during model merging, ensuring that essential information is maintained. Model merging methods of this type include the following.

\textbf{TIES-MERGING: } This method focuses on minimizing parameter redundancy and resolving sign conflicts, two primary sources of performance degradation in model merging. TIES-MERGING reduces interference in model merging by selectively keeping important parameters and resolving sign conflicts. First, it drops low-magnitude parameters, resetting them to pre-trained values while preserving only the most significant ones. Then, it resolves conflicting parameter signs by selecting the dominant direction across models. Finally, it merges only the aligned parameters, preventing destructive interference. This method improves merging quality by ensuring that only the most relevant and consistent updates are retained~\cite{yadav2024ties}.
\begin{figure}[htbp]
    \centering
    \includegraphics[width=\linewidth]{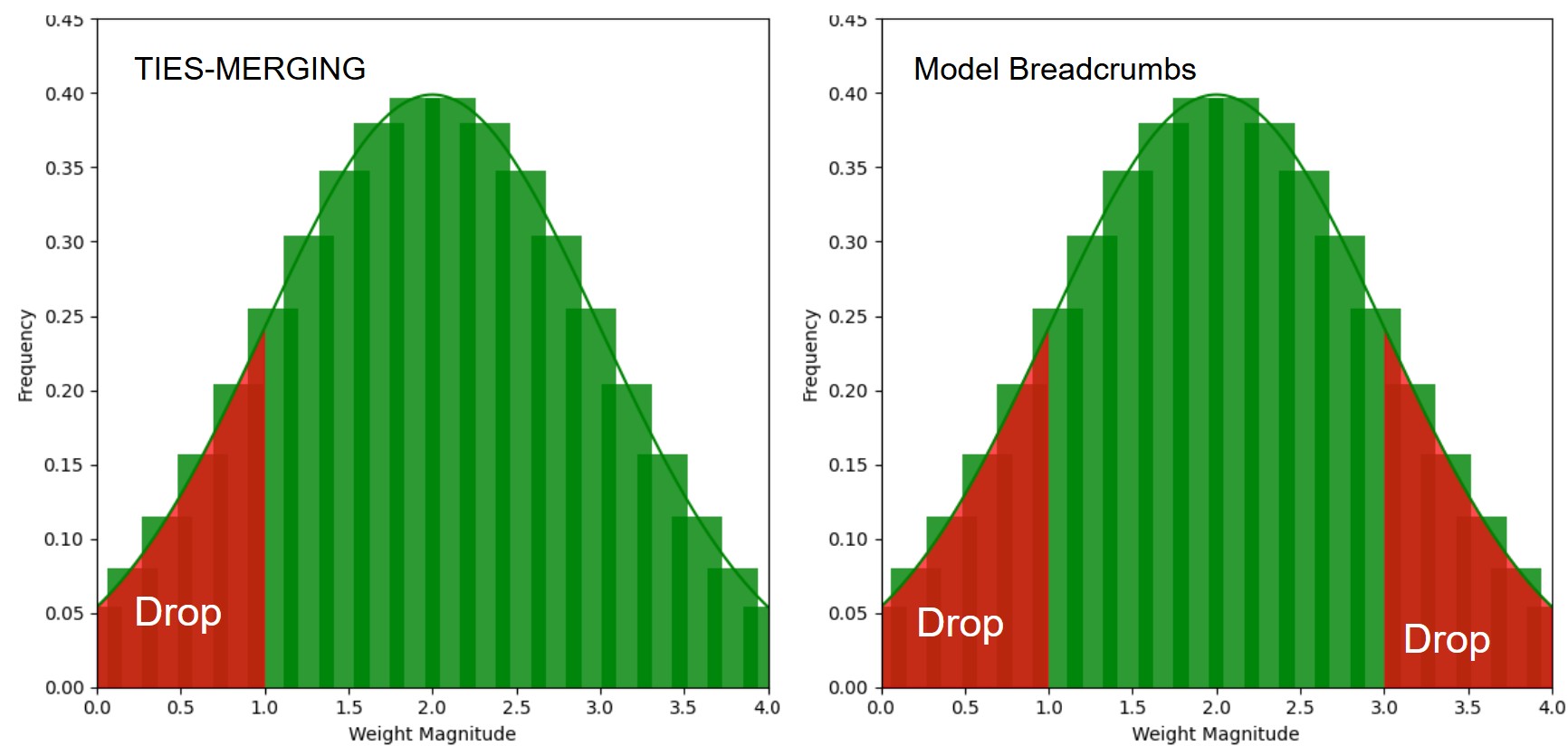} 
    \caption{A schematic representation illustrating the drop principles in two model merging methods: TIES-MERGING and Model Breadcrumbs. TIES-MERGING focuses on dropping low-magnitude weights, while Model Breadcrumbs applies predefined thresholds to balance sparsity and parameter retention.} 
    \label{fig:2} 
\end{figure}

\begin{table*}[t!]
\centering
\caption{Characteristics of Different Model Merging Techniques}
\small
\newcolumntype{C}[1]{>{\centering\arraybackslash}m{#1}}

\begin{tabular}{|C{4.5cm}|C{1.2cm}|C{2.0cm}|C{1.5cm}|C{1.5cm}|C{2cm}|C{1.2cm}|}
\hline
\textbf{Method} & \textbf{Drop} & \textbf{ Drop criteria} & \textbf{Rescale} & \textbf{Direction Selection} & \textbf{Requires partial Training Data}&\textbf{Layerwise Uniform} \\ \hline

FREE-Merging~\cite{wang2024localizing} & $\checkmark$ & 
large features
& $\times$ & total sum & $\times$ &$\checkmark$\\ \hline

TIES-Merging~\cite{yadav2024ties} & $\checkmark$ & large weight & $\times$ & total sum & $\times$ &$\checkmark$\\ \hline

DARE~\cite{yu2024language} & $\checkmark$ & random  & \( \dfrac{1}{1-p} \) & total sum& $\times$&$\checkmark$\\ \hline

DAREx~\cite{deng2024dare} & $\checkmark$ & random & \( q \) & total sum & $\checkmark$ &$\times$\\ \hline

DELLA-Merging~\cite{deep2024della} & $\checkmark$ & weighted drop & \( \dfrac{1}{1-p} \) & total sum & $\checkmark$ &$\checkmark$ \\ \hline

EMR-Merging~\cite{huang2024emr} & Mask & 
direction
& $\checkmark$ & max weight & $\times$ &$\checkmark$\\ \hline

WIDEN~\cite{yu2024extend} & $\times$ & $\times$ & direction and weight & Calculating & $\checkmark$ &$\checkmark$ \\ \hline

Model Breadcrumbs~\cite{davari2024model} & $\checkmark$ & remove outliers & $\times$& total sum &$\checkmark$&$\checkmark$ \\ \hline

PCB-Merging~\cite{du2024parameter} & $\checkmark$ & Intra-task and Inter-task  & $\checkmark$ & $\times$&$\checkmark$ &$\times$\\ \hline

SurgeryV2~\cite{yang2024surgeryv2} & $\times$ & $\times$  & $\times$ & $\times$&$\checkmark$ &$\times$\\ \hline

Zipit!~\cite{stoica2023zipit}& $\times$ &$\times$  & $\times$ & $\times$&$\checkmark$ &$\times$\\ \hline

Localize-and-Stitch~\cite{he2024localize} & $\checkmark$ & output results  & $\times$ & total sum&$\checkmark$ &$\times$\\ \hline

Fisher~\cite{matena2022merging} & $\checkmark$ & $\times$  & $\times$ & $\times$&$\checkmark$ &$\times$\\ \hline

AdaMerging~\cite{yang2023adamerging} & $\times$ & $\times$  & $\times$ & $\times$&$\checkmark$ &$\times$\\ \hline

\end{tabular}
\label{tab:methods}
\end{table*}


\textbf{Model Breadcrumbs:} 
  Unlike TIES-Merging, which resolves sign conflicts and trims low-magnitude parameters to address interference (Figure~\ref{fig:2}), Model Breadcrumbs employs a dual masking strategy that simultaneously removes large outliers and small perturbations in task vectors. This ensures a more balanced and noise-resistant parameter selection~\cite{davari2024model}. In contrast to DARE, which randomly drops delta parameters with a fixed probability, Model Breadcrumbs applies a deterministic, layer-wise masking process that accounts for task vector distributions, allowing for finer control over parameter adjustments. Compared to DELLA-Merging’s stochastic pruning strategy, MAGPRUNE, which relies on magnitude-based probability sampling, Model Breadcrumbs adopts a more direct filtering mechanism that eliminates extreme values and maintains task relevance through robust aggregation. Furthermore, Model Breadcrumbs emphasizes layer-specific processing over global operations, enhancing its adaptability to diverse tasks. These distinctions make Model Breadcrumbs particularly effective in reducing noise accumulation and achieving consistent performance across multi-task scenarios, while requiring less hyperparameter tuning than prior methods.

\subsubsection{Decomposing weight type}

This type of model merging not only considers weight magnitudes to address conflicts and interference but also decomposes the weights to analyze both their magnitude and direction. This approach further mitigates conflicts between models. Below, we introduce representative methods.

\textbf{PCB-Masks:}
This method introduces task-specific binary masks to improve model merging and compression.
Binary masks are generated for each task using task vectors from fine-tuned models, highlighting parameters important for that task while ignoring irrelevant ones.
 During merging, the method eliminates "selfish weights" (important to one task only) and "catastrophic weights" (irrelevant to all tasks), preserving only shared parameters that benefit multiple tasks~\cite{wang2024localizing}.

\textbf{EMR-MERGING:} 
The EMR-Merging method enables tuning-free model merging by selecting the maximum absolute value of each parameter while preserving the dominant sign direction, reducing interference. It then applies task-specific masks to filter conflicting signs and rescalers to adjust parameter magnitudes. During inference, these modulators adapt the merged model to different tasks, achieving high accuracy across vision, NLP, and multi-modal models without additional training~\cite{huang2024emr}.

\textbf{WIDEN(Weight Disentanglement ):} 
A novel approach to extending model merging techniques beyond fine-tuned (FT) models to also include pre-trained (PT) models. The key idea behind WIDEN is to disentangle model weights into two components: magnitude and direction. By quantifying the divergence of these components from the backbone model, WIDEN can automatically determine the importance of each model in the merging process, eliminating the need for manually assigned scaling factors~\cite{yu2024extend}. Additionally, it employs a Softmax-based score calibration to adaptively balance the contributions of different models, ensuring that the merged model retains and optimally integrates their abilities (Table\ref{tab:methods}).

\textbf{FREE-Merging:} It is a novel model merging approach that leverages Fourier transform-based filtering and lightweight expert modules~\cite{zheng2024free}. It mitigates task conflicts by applying high-pass filtering in the frequency domain, removing low-frequency signals that reduce generalization while preserving essential model structures.

To compensate for information loss, it introduces lightweight task-specific experts that selectively store crucial parameters using rescaling and extraction techniques. These experts are dynamically integrated via a router, ensuring efficiency with minimal storage and computation. This two-stage strategy balances merging efficiency, performance, and deployment costs, making it ideal for multi-task learning with limited resources.



\subsection{Activation-based Pruning Merging type}

Relying solely on weight magnitude for pruning is insufficient, as it does not fully capture a model's functional contributions. A common alternative is pruning based on model activations or output responses, which helps reduce conflicts and improve merging effectiveness. By selectively retaining important neurons and representations, this approach enhances model performance. Representative methods include:

\textbf{ZipIt!:} This method merges models trained on disjoint tasks by identifying shared feature activations and selectively "zipping" them together. It introduces a partial merging strategy, allowing models to retain distinct task-specific heads while merging up to a certain layer~\cite{stoica2023zipit}.

\textbf{SurgeryV2:} Addresses representation bias in model merging by performing layer-wise representation surgery. Instead of directly averaging weights, it aligns activations between expert models and the merged model, progressively refining feature representations to reduce task interference~\cite{yang2024surgeryv2}.

\textbf{Localize-and-Stitch:} A localized merging approach that first identifies sparse task-specific activation regions and stitches only these into the final model. By selecting highly activated neurons, it minimizes task conflicts and retains essential task-specific knowledge~\cite{he2024localize}.

These methods improve model merging efficiency by pruning low-importance activations, ensuring better generalization across tasks while reducing parameter redundancy.

\textbf{AdaMerging:} It assigns different sparsity rates to different layers, pruning less important layers more aggressively while preserving critical layers~\cite{yang2023adamerging}. This ensures that merging does not indiscriminately combine task-specific features, reducing task interference. Instead of relying solely on weight magnitudes, AdaMerging minimizes entropy to ensure more deterministic outputs during merging, making it an activation-aware approach.

\subsection{Hybrid Pruning Merging type}

Some model merging approaches take a hybrid pruning strategy, considering both activation responses and weight importance to determine which parameters should be retained before merging. These methods identify important parameters by evaluating how weights influence activations across tasks, ensuring that only the most essential components are preserved. Parameters deemed less critical are pruned, reducing redundancy and mitigating task conflicts before performing the merging process.

Two notable methods in this category are MACL (Model Sensitivity Aware Continual Learning) and PCB-Merging (Parameter Competition Balancing for Model Merging). While both leverage activation and weight-based information, they differ in their specific strategies. MACL focuses on minimizing parameter sensitivity, ensuring that small updates do not drastically alter activations, thereby maintaining knowledge stability across continual learning tasks~\cite{wangmodel}. PCB-Merging, on the other hand, introduces parameter competition balancing, dynamically adjusting parameter scaling through intra-task and inter-task balancing mechanisms. These differences allow MACL to excel in maintaining long-term task stability, while PCB-Merging optimizes for multi-task integration by resolving conflicts at the parameter level~\cite{du2024parameter}.

\subsection{Optimization merging type}
Optimization-based merging methods, such as Fisher-weighted averaging and Regression Mean (RegMean), offer precise but computationally expensive solutions for combining fine-tuned models~\cite{matena2022merging,jin2022dataless}. Instead of relying on simple heuristics, these methods optimize parameter combinations to improve accuracy and generalization.

Fisher-weighted averaging leverages the Fisher information matrix to assign importance to each parameter during merging~\cite{matena2022merging}. It prioritizes parameters with higher sensitivity to model predictions, ensuring a more informed merging process. However, computing the Fisher information matrix involves high computational costs, making it less practical for large-scale models.

RegMean treats model merging as a regression problem, aiming to minimize the discrepancy between the merged and individual models~\cite{jin2022dataless}. Assuming a linear relationship between parameters and inputs, it provides a closed-form solution that optimally balances model weights. This method is efficient but may not capture complex nonlinear dependencies.

Other notable optimization-based approaches include AdaMerging\cite{yang2023adamerging}, which adapts merging weights dynamically. These methods aim to enhance accuracy, efficiency, and generalization while managing computational constraints.

\subsection{LoRA merging type}

With the rapid rise of large language models, an increasing number of fine-tuned models, particularly LoRA fine-tuned models, have been open-sourced. To accommodate different downstream tasks, the merging of LoRA models has developed rapidly in recent years. Although merging LoRA models is fundamentally similar to conventional model merging, many standard computationally intensive methods are no longer feasible due to the nature of large language models. As a result, it is necessary to develop merging techniques specifically tailored to the characteristics of LoRA models. In this paper, we categorize these techniques as the LoRA model merging class. Among these approaches, several have gained significant influence, including Twin Merging~\cite{lu2024twin},  OpenMoE~\cite{xue2024openmoe}, and LoraHub~\cite{huang2023lorahub}. Additionally, most of these methods integrate with MoE techniques, forming distinctive hybrid approaches. In the following, we take LoraHub as an example to illustrate the characteristics of this class of methods.
LoraHub is a method that dynamically merges multiple LoRA modules to achieve efficient cross-task generalization without requiring additional parameters or gradients. It automatically composes task-specific LoRA modules and optimizes their combination using a gradient-free approach, allowing adaptation to new tasks with just a few examples.

By leveraging pre-trained LoRA modules, LoraHub eliminates the need for full fine-tuning, significantly reducing computational costs. It achieves performance close to in-context learning while using far fewer tokens per example. This makes it a lightweight, modular, and scalable approach for adapting large language models to diverse tasks.

\section{Challenges and Future Perspectives of
Model Merging}

Model merging, while gaining traction and demonstrating significant potential, still encounters several key challenges that must be addressed to achieve broader adoption and improved efficiency.

Firstly, as the number of tasks increases, merged models often underperform compared to independent expert models. Maintaining consistent performance across diverse tasks without extensive task-specific tuning remains a significant hurdle. Moreover, merging models trained on different tasks or domains can result in conflicts where knowledge from one model interferes with another. This issue is particularly pronounced in scenarios like multi-task learning and continual learning, where tasks often have distinct requirements.

Additionally, the lack of comprehensive theoretical frameworks for model merging limits the ability to predict and guarantee performance. Many current methods rely heavily on heuristic or empirical strategies, leaving room for improvement through deeper theoretical exploration.

Lastly, identifying optimal merging strategies, such as determining appropriate weight coefficients for models or parameters, is difficult. Fine-grained optimization approaches often come with high computational and resource demands, making them less feasible in practice.

Addressing these challenges will require innovative methodologies, advanced computational tools, and a stronger theoretical foundation to support the efficient and reliable implementation of model merging across diverse applications.

Combining model compression with model merging represents one of the promising future directions for advancing model merging techniques~\cite{wang2024localizing,lu2024twin}. Model compression, which involves reducing the size and complexity of a model while preserving its performance, helps mitigate interference between models during the merging process. By streamlining individual models before merging, compression can enhance compatibility and lead to more effective and seamless integration of model parameters, ultimately improving the overall merging outcomes.

Apart from leveraging model merging techniques, task merging or classification can also be explored as potential approaches. For instance, the Disperse-Then-Merge (DTM) method provides an innovative framework for addressing alignment tax in supervised fine-tuning of large language models~\cite{fu2024disperse}. This method tackles the issue of data biases by dividing the instruction-following dataset into several clusters, training separate sub-models on these clusters, and subsequently merging the sub-models into a single model. By doing so, DTM effectively distributes and mitigates dataset-specific biases, maintaining the model's instruction-following capacity while reducing the detrimental effects of such biases on knowledge and reasoning benchmarks. This simple yet efficient approach demonstrates the promise of task-level separation and targeted training strategies to enhance model performance in multi-task settings.

In addition to operations on weights and activations, future research could benefit from a deeper exploration of network architecture design as a pivotal factor in enhancing model performance during merging and continual learning. As highlighted in recent work, different neural network structures exhibit significantly varying degrees of resilience to catastrophic forgetting, a core challenge in continual learning ~\cite{lu2024revisiting}. Tailoring merging methods to align with these structural features could lead to more effective adaptation and knowledge retention across tasks. By leveraging insights into the unique structural properties of models, it becomes possible to optimize merging strategies not just for parameter efficiency, but also for preserving task-specific knowledge in multi-task settings.

This direction underscores the importance of designing architecture-aware methods in model merging, where decisions about how and where to merge layers or weights could be dynamically adjusted based on architectural traits. Such an approach could unlock new possibilities for mitigating catastrophic forgetting while maintaining model scalability and generalization.

In addition to the integration of model compression and model merging as a significant future trend, several other methodologies hold substantial development potential. These include the application of dual-space constraints—encompassing both weight space and activation space—to enhance model performance and generalization. Furthermore, decomposing model gradients into their directional and magnitude components can provide deeper insights into optimization processes. Lastly, employing clustering techniques to select appropriate experts within models can lead to more efficient and specialized architectures. Collectively, these approaches offer promising avenues for advancing model efficiency and effectiveness.

\section{Conclusion}
Model merging is a straightforward and effective technique for enhancing models by combining multiple models to achieve diverse capabilities. In this paper, we classified the current common merging methods, compared the differences among methods within the same category, and summarized their commonalities. We also pointed out existing problems in the field of model merging and discussed future trends.

\newpage

\appendix



\bibliographystyle{named}
\bibliography{ijcai25}

\end{document}